\documentclass[letterpaper]{article} % DO NOT CHANGE THIS
\usepackage[submission]{aaai24}  % DO NOT CHANGE THIS
\usepackage{times}  % DO NOT CHANGE THIS
\usepackage{helvet}  % DO NOT CHANGE THIS
\usepackage{courier}  % DO NOT CHANGE THIS
\usepackage[hyphens]{url}  % DO NOT CHANGE THIS
\usepackage{graphicx} % DO NOT CHANGE THIS
\usepackage{natbib}  % DO NOT CHANGE THIS AND DO NOT ADD ANY OPTIONS TO IT
\usepackage{caption} % DO NOT CHANGE THIS AND DO NOT ADD ANY OPTIONS TO IT
\usepackage{algorithm}
\usepackage{algorithmic}

\usepackage{algorithm}
\usepackage{algorithmic}
\usepackage{adjustbox} %调整表格大小
\usepackage{booktabs}
\usepackage{diagbox}
\usepackage{multirow}
\usepackage{amsmath}
\usepackage{amsfonts}
\usepackage{amssymb}
\usepackage{makecell}
\usepackage{subfigure}
\usepackage{colortbl}
\usepackage{xcolor}
\usepackage[switch]{lineno}

\usepackage{newfloat}
\usepackage{listings}
\usepackage{bbding}

\DeclareCaptionStyle{ruled}{labelfont=normalfont,labelsep=colon,strut=off} % DO NOT CHANGE THIS
\floatstyle{ruled}
\newfloat{listing}{tb}{lst}{}
\floatname{listing}{Listing}
\title{Exploring Sparse Visual Prompt for Domain Adaptive Dense Prediction}

\author{Senqiao Yang\textsuperscript{\rm 1,2}\thanks{Equal contribution: yangsenqiao.ai@gmail.com, \\ $^{\dagger}$Project leader: jiamingliu@stu.pku.edu.cn, \\ \textsuperscript{\Envelope} Corresponding author: shanghang@pku.edu.cn}, 
Jiarui Wu\textsuperscript{\rm 1,3*},
Jiaming Liu \textsuperscript{\rm 1*$^{\dagger}$}, 
 Xiaoqi Li\textsuperscript{\rm 1},
Qizhe Zhang\textsuperscript{\rm 1},\\
Mingjie Pan\textsuperscript{\rm 1},
Yulu Gan\textsuperscript{\rm 1},
Zehui Chen \textsuperscript{\rm 4},
Shanghang Zhang\textsuperscript{\rm 1}~\textsuperscript{\Envelope}\\
\textsuperscript{\rm 1}Peking University, \textsuperscript{\rm 2}Harbin Institute of Technology, Shenzhen,\\ \textsuperscript{\rm 3} Beihang University, \textsuperscript{\rm 4} University of Science and Technology of China
}

\usepackage{bibentry}
\usepackage{hyperref}
\begin{document}

\maketitle

\begin{abstract}
The visual prompts have provided an efficient manner in addressing visual cross-domain problems. In previous works, \cite{gan2022decorate} first introduces domain prompts to tackle the classification Test-Time Adaptation (TTA) problem by placing image-level prompts on the input and fine-tuning prompts for each target domain. However, since the image-level prompts mask out continuous spatial details in the prompt-allocated region, it will suffer from inaccurate contextual information and limited domain knowledge extraction, particularly when dealing with dense prediction TTA problems. To overcome these challenges, we propose a novel Sparse Visual Domain Prompts (SVDP) approach, which applies minimal trainable parameters (e.g., 0.1\%) to pixels across the entire image and reserves more spatial information of the input. To better apply SVDP in extracting domain-specific knowledge, we introduce the Domain Prompt Placement (DPP) method to adaptively allocates trainable parameters of SVDP on the pixels with large distribution shifts. Furthermore, recognizing that each target domain sample exhibits a unique domain shift, we design Domain Prompt Updating (DPU) strategy to optimize prompt parameters differently for each sample, facilitating efficient adaptation to the target domain. Extensive experiments were conducted on widely-used TTA and continual TTA benchmarks, and our proposed method achieves state-of-the-art performance in both semantic segmentation and depth estimation tasks. Project page: \href{http://Senqiaoyang.com/project/svdp}{Senqiaoyang.com/project/svdp}
\end{abstract}

\section{Introduction}

\begin{figure}[t]
\includegraphics[width=0.47\textwidth]{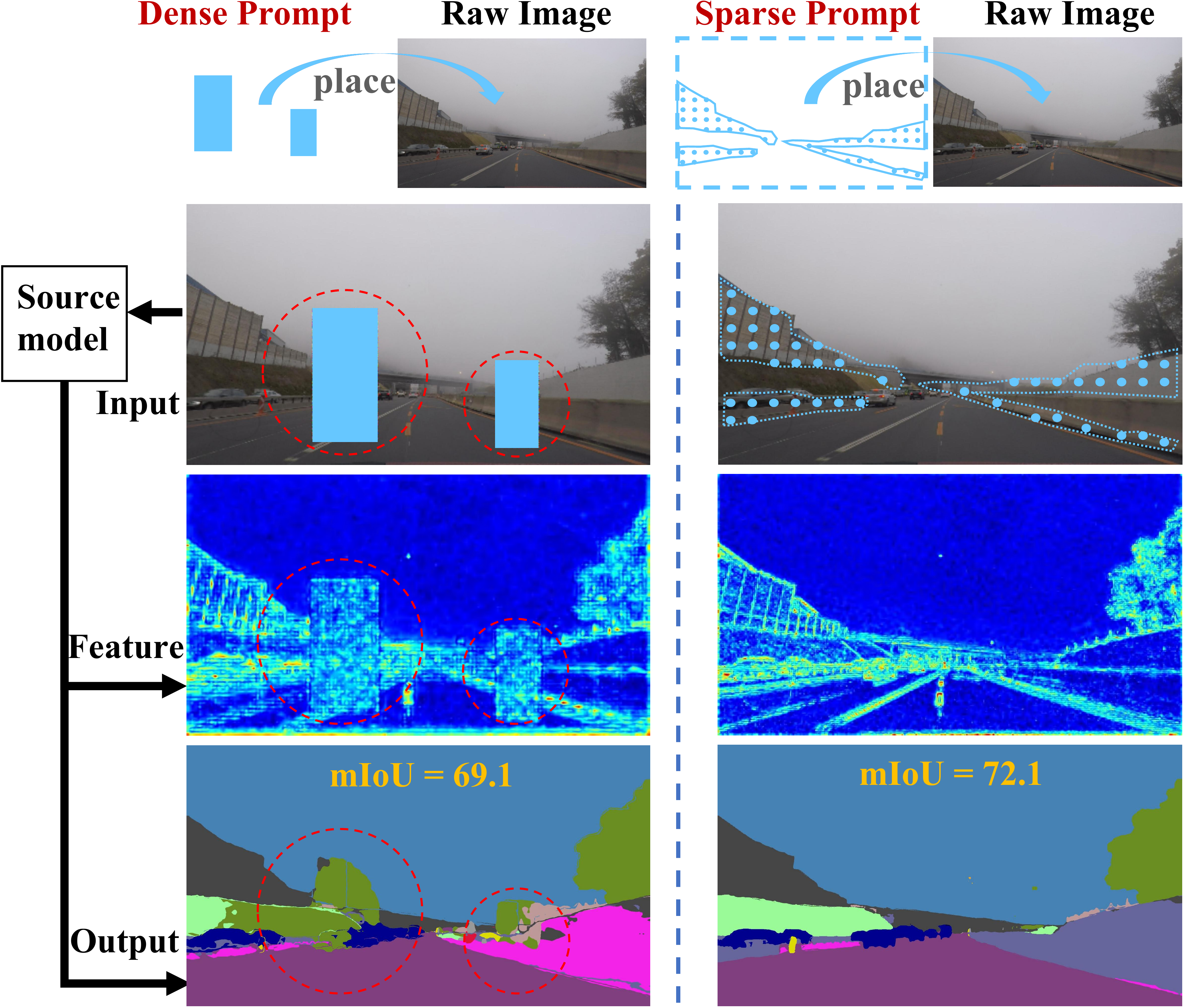}
\vspace{-0.4cm}
\centering
\caption{
\textbf{The motivation and main idea of our method.} 
\textcolor{red}{(a)} Previous dense visual domain prompts (VDP) mask out consecutive spatial details in the placed regions as shown in red circles. In dense prediction DA problems, applying dense VDP will lead to inaccurate context information extraction and severe performance degradation. \textcolor{red}{(b)} We introduce Sparse Visual Domain Prompts (SVDP), which are tailored for addressing the occlusion problem of pixel-wise information and can better extract local domain knowledge for cross-domain learning. Though the parameters of SVDP are less than VDP, SVDP achieves better semantic segmentation performance in the Test Time Adaptation.}
\label{fig:intro}
\vspace{-0.5cm}
\end{figure}

Deep neural networks can achieve promising performance if test data is of the same distribution as the training data. However, it is not the common case in real-world scenarios \cite{Radosavovic2022}, which contain diverse and disparate domains. When applying a pre-trained model in real-world tasks, the domain gap commonly exists \cite{sakaridis2021acdc}, leading to significant performance degradation on target data. Though we can manually collect labeled data for each real-world target domain, it is laborious and time-consuming~\cite{chen2022multi}. To this end, the domain adaptation (DA) methods are introduced and have drawn growing attention in the community.

While DA extensively investigates to address distribution shifts, its typical assumption involves access to raw source data. However, in real-world scenarios, raw data often cannot be publicly accessible due to data protection regulations. Meanwhile, traditional DA methods present resource-intensive backward computation, leading to high training costs \cite{ganin2015unsupervised}. To address this, Test-time adaptation (TTA) \cite{liang2023comprehensive} is gained significant attention, which tackles distribution shifts at test time with only unlabeled test data streams. Prior TTA studies \cite{DequanWangetal2021,Wangetal2022, chen2022contrastive, goyal2022test} predominantly focus on model-based adaptation, utilizing model parameters to fit target domain knowledge.

To better solve the TTA problem, motivated by the recent advances of prompting in NLP \cite{li2021prefix, liu2023pre}, VDP \cite{gan2022decorate} first introduces a prompt-based method to tackle the classification TTA problem. It employs image-level prompts to enhance domain transfer efficiency and effectiveness. Specifically, it randomly places the dense prompt on the input image and fine-tunes it to extract target domain knowledge. However, this prompt-based technique encounters limitations when applied to dense prediction tasks such as semantic segmentation and depth estimation TTA. Specifically, the dense prompts obscure continuous spatial information in the allocated regions, as illustrated in Figure 1 (a). This occlusion introduced by prompts leads to incomplete semantic knowledge representation, thereby negatively impacting the quality of segmentation maps. Simultaneously, the occluded details within corresponding features impede the extraction of adequate domain knowledge during cross-domain learning.
 
To this end, as shown in Fig.1 (b), we propose a novel Sparse Visual Domain Prompts (SVDP) approach for effectively extracting target domain knowledge, specially designed to combat domain shifts in dense prediction tasks. By introducing sparse prompts, which apply minimal trainable parameters (e.g., 0.1\%) to pixels across the entire image, more spatial information from the input is preserved. Furthermore, the semantic information can be extracted sufficiently (shown in line 2), leading to noticeable improvements in segmentation outcomes (shown in line 3). 
In order to better apply SVDP in the pixel-wise TTA task, we propose the Domain Prompt Placement (DPP) to adaptively allocate trainable parameters of SVDP on the pixel with large distribution shifts. In this way, SVDP excels at extracting local domain knowledge, facilitating the transfer of pixel-wise data distribution from the source to the target domain. Furthermore, recognizing that each target domain sample exhibits a unique domain shift, we design a Domain Prompt Updating (DPU) method to optimize prompt parameters efficiently during the TTA process. Specifically, based on the extent of the domain gap observed in target domain samples, we employ varying weights to update the visual prompts.
It's worth noting that we are the pioneers in designing specific strategies for pixel-level placement and image-level optimization in vision prompt learning, which work in synergy to address domain shifts in dense prediction TTA tasks.

Since data privacy and transmission limit access to source data in the real world, we evaluate our method on semantic segmentation and depth estimation source-free adaptation settings, including online TTA~\cite{Liangetal2020} and Continual Test-Time Adaptation~\cite{Wangetal2022} (CTTA). Our proposed approach demonstrates superior performance compared to most state-of-the-art (SOTA) methods across three benchmarks, covering Cityscapes to ACDC~\cite{sakaridis2021acdc} and KITTI \cite{geiger2012we} to DrivingStereo \cite{yang2019drivingstereo}. 
The main contributions are shown as follows:

1) We are the first to introduce the visual prompt approach to the dense prediction TTA problem. We propose a novel Sparse Visual Domain Prompts (SVDP) approach to better extract local domain knowledge and transfer pixel-wise data distribution from the source to the target domain.

2) In order to efficiently apply SVDP in pixel-wise TTA tasks, we propose the Domain Prompt Placement~(DPP) method to adaptively allocate trainable parameters in SVDP based on the degree of distribution shift at the pixel level. And Domain Prompt Updating~(DPU) is designed to optimize prompt parameters differently for each sample, facilitating efficient adaptation on target domains.

3) We conduct extensive experiments to evaluate the effectiveness of our method, which outperforms most SOTA methods on four TTA and two CTTA benchmarks, covering semantic segmentation and depth estimation tasks.

\section{Related Work}
\subsection{Test-time adaptation}        
\textbf{Test-time adaptation (TTA)},~\cite{Boudiafetal2023, Kunduetal2022, Liangetal2020, ShiqiYangetal2021}, aims to adapt a source model to an unknown target domain distribution without using any source domain data. 
Recent research has focused on using self-training or entropy regularization to fine-tune the source model. Specifically, 
SHOT~\cite{Liangetal2020} optimizes only the feature extractor using information maximization and pseudo labeling. AdaContrast~\cite{Chenetal2022} also uses pseudo labeling for TTA, but introduces self-supervised contrastive learning to enhance performance. In addition to model-level adaptation, ~\cite{Boudiafetal2022} adjusts the output distribution to address this problem. Tent~\cite{DequanWangetal2021} updates the training parameters in the batch normalization layers by entropy minimization. Recent works \cite{niu2023towards,
yuan2023robust} follow Tent to continually explore the robustness of normalization layers in the TTA process.
While the aforementioned works primarily focus on classification tasks, there has been a recent surge of interest in performing TTA on dense prediction tasks~\cite{Shinetal022, Songetal2022, Zhangetal2021}. 

\textbf{Continual Test-Time Adaptation (CTTA)} is a scenario in which the target domain is not static, increasing challenges for traditional TTA methods ~\cite{Wangetal2022}. ~\cite{Wangetal2022} serves as the first approach to tackle this task, using a combination of bi-average pseudo labels and stochastic weight reset. While ~\cite{Wangetal2022,song2023ecotta} addresses the continual shifts at the model level, ~\cite{gan2022decorate} leverages visual domain prompts to address the problem in the classification task at the input level for the first time. In this paper, we evaluate our approach on both TTA and CTTA with a specific focus on the dense prediction task.

\subsection{Prompt learning}
\textbf{Visual prompts} are inspired by their counterparts~\cite{PengfeiLiu2021PretrainPA} which are used in natural language processing (NLP). Language prompts are presented as text instructions to improve the pre-trained language model's understanding of downstream tasks~\cite{Brownetal2020}.
Recently, researchers have attempted to discard text encoders and use prompts directly for visual tasks. ~\cite{Bahngetal2022} employs visual prompts to pad input images, enabling pre-trained models to adapt to new tasks. Rather than fine-tuning the entire model, VPT~\cite{Conderetal2022, MenglinJia2022VisualPT, Sandleretal2022, ZifengWangetal2021} inserts prompts into image or feature-level patches to adapt Transformer-based models. While these approaches all utilize opaque-block prompts, such prompts can cause performance degradation in dense prediction tasks. 
\textbf{Domain prompts} are first introduced in DAPL~\cite{Geetal2022}, which proposes a novel prompt learning paradigm for unsupervised domain adaptation (UDA). Embedding domain information using prompts can minimize the cost of fine-tuning and enable efficient domain adaptation. Recognizing the potential of prompt learning for UDA, MPA~\cite{chen2022multi} proposes multi-prompt alignment for multi-source UDA. DePT~\cite{Gaoetal2022} combines domain prompts with a hierarchical self-supervised regularization for TTA, which aims to solve the error accumulation problem in self-training. ~\cite{gan2022decorate} further divides domain prompts into domain-specific ones and domain-agnostic ones to address the more challenging CTTA task. However, these studies mainly focus on simple classification DA tasks. Our method, for the first time, applies sparse domain prompts to dense prediction DA tasks. Besides, we are the first to design specific placement and updating strategies for the domain prompt method, which helps to jointly ease the domain shift.

\begin{figure*}[ht]
\centering
\includegraphics[width=\linewidth]{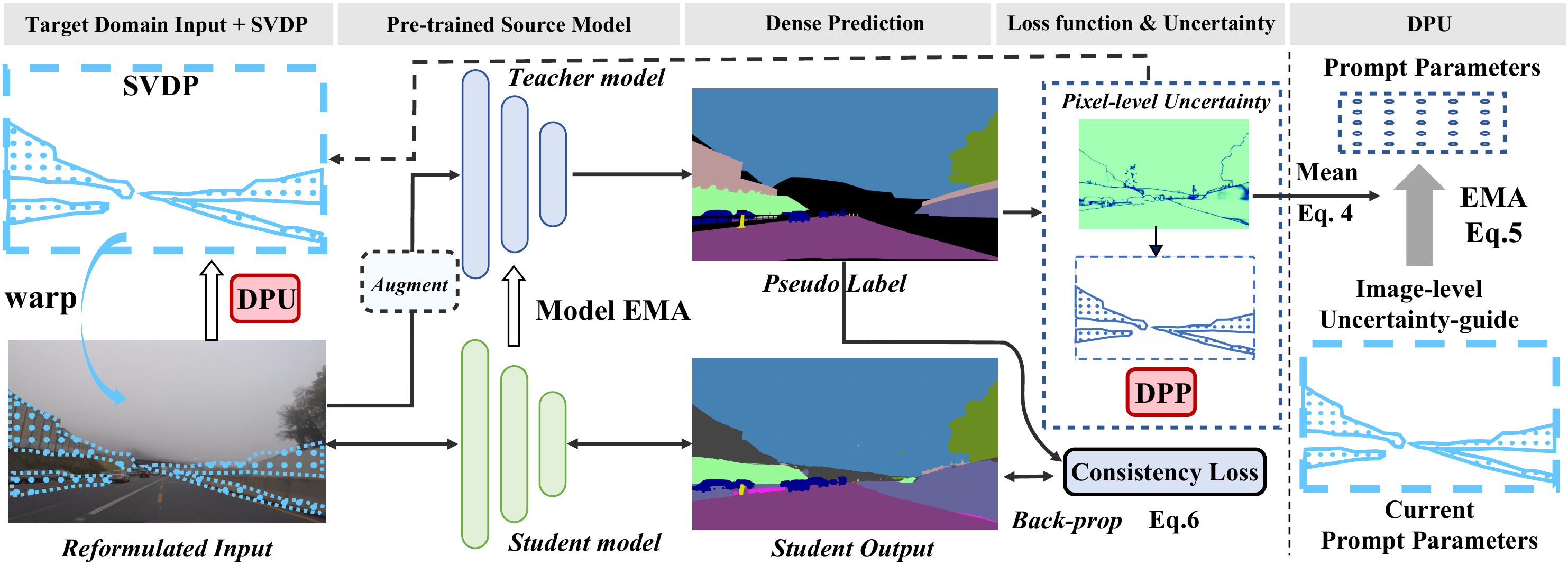}
% \vspace{-0.45cm}
\caption{\textbf{The overall framework.} \textbf{Left:} We warp the SVDP into the image and place prompt parameters on the selected pixel by the Domain Prompt Placement (DPP) method. The reformulated image serves as the input of the teacher and student model.  
We obtain the uncertainty map as described in Eq. \eqref{eq:mc} through the teacher model. The uncertainty map is used to evaluate the degree of pixel-level distribution shift. 
SVDP adopts consistency loss (Eq. \eqref{eq:loss}) and exponential moving average (EMA) as the optimization strategies.
\textbf{Right:} Domain Prompt Updating (DPU). Based on the image-level uncertainty value, we adopt different EMA weights to realize stable updating of SVDP parameters, facilitating efficient adaptation to the target domain.}
\label{fig:framework} 
% \vspace{-0.2cm}
\end{figure*}

\section{Method}

\subsection{Preliminaries}

\textbf{Test Time Adaptation (TTA)} ~\cite{liang2023comprehensive} aims at adapting a pre-trained model with parameters trained on the source data $(\mathcal{X}_S$, $\mathcal{Y}_S)$ to multiple unlabeled target data distribution $\mathcal{X}_{T_1},\mathcal{X}_{T_2}, \dots, \mathcal{X}_{T_n}$ at inference time. The entire process can not access any source domain data and can only access target domain data once. $\mathcal{X}_{T_i} = \{x_i^T\}_{i=1}^{N_t}$, where $N_t$ denotes the scale of the target domain. The upcoming target domain can be a single one (TTA) or multiple continually changing distributions (CTTA), the latter of which is a more realistic setting that requires the model to achieve stability while preserving plasticity. 

\textbf{Domain Prompt.} Inspired by language prompt in NLP, ~\cite{gan2022decorate} first introduces visual domain prompt (VDP) 
serving as a reminder to continually adapt to the target domain for the classification task, which aims to extract target domain-specific knowledge. Specifically, VDP ($\textbf{p}$) are dense learnable parameters added to the input image. 
\begin{equation}
\widetilde {\textbf{x}} = \textbf{x} + \textbf{p}
\label{eq:dense}
\end{equation}
where x represents the original input image. The reformulated image ${\widetilde {\textbf{x}}}$ will serve as the input for our model instead.

\subsection{Motivation}
\label{sec:3.2}

\textbf{Sparse Visual Domain Prompt.} Traditional visual prompts \cite{jia2022visual} are deployed on the image or feature level to realize fine-tuning by updating a small number of prompt parameters. Recent works \cite{gan2022decorate, gao2022visual} explore visual prompts in classification DA problems, which extract domain knowledge for the target domain and transfer data distribution from the source to the target domain. 
However, DePT \cite{gao2022visual} concatenates the domain prompts with class token and image tokens to the input of transformer layers, which neglect the local domain knowledge extraction.
Meanwhile, VDP \cite{gan2022decorate} randomly set the locations of dense prompts on the input image, masking out continuous spatial details in prompt allocated regions. 
Different from classification cross-domain learning, dense prediction DA not only requires global domain knowledge but also relies on extracting intact local domain knowledge.  
As shown in Fig.\ref{fig:intro}(a), partial spatial information deficiency caused by dense prompts will lead to inaccurate contextual information and negative effects on target domain knowledge extraction. 
This observation motivates us to propose a novel Sparse Visual Domain Prompts (SVDP), which is tailored for pixel-wise prediction DA tasks. It inserts minimal trainable parameters into pixels across the entire image and reserves more spatial information.

\textbf{Domain Prompt Placement.} Previous work~\cite{gan2022decorate,gao2022visual} randomly put the prompts on the target domain image to extract global domain knowledge. Specifically, it may set prompts on regions with trivial domain shifts, hindering the extraction of local domain knowledge. Especially in the source-free TTA setting, we can only access target domain data once, which makes the efficiency of transfer learning crucial. Therefore, we propose Domain Prompt Placement (DPP) which efficiently extracts more domain-specific knowledge and addresses local domain shift. Specifically, we measure the degree of domain gap by general uncertainty scheme~\cite{gal2016dropout, guan2021uncertainty, roy2022uncertainty,gan2022cloud} and tactfully place trainable parameters of SVDP on the pixel with large distribution shifts.

\textbf{Domain Prompt Updating.}
The amount of prompt parameters is minimal which brings the challenge of fully learning target domain knowledge during TTA process. Meanwhile, the degree of domain shift is not only various on pixels within the image but also on each target domain test sample. It thus motivates us to update prompt parameters differently for each target sample. Therefore, we design a Domain Prompt Updating (DPU) which efficiently optimizes prompt parameters to fit in target domain distribution. Specifically, we adopt the same uncertainty scheme to measure the degree of domain shift for each target sample.
According to the degree, we update prompt parameters for the individual sample with different updating weights.

\subsection{Sparse visual domain prompt} 
\label{sec:3.3}
SVDP maintains the same resolution as the input image ($\textbf{p} \in \mathbb{R}^{H \times W \times 3}$), it only masks out original information by minimal discrete trainable parameters (e.g. $0.1\%$) on the pixels with large domain shifts. Compared with the previous dense visual prompt, SVDP preserves more contextual information and possesses the capacity to capture local domain knowledge through pixel-wise prompt parameters. The overall framework of our method is shown in Fig .\ref{fig:framework}, and the specially designed prompt Placement and Updating methods are introduced in the following.

\subsection{Domain prompt placing}
\label{sec:3.4}

\begin{figure}[t]
\includegraphics[width=0.45\textwidth]{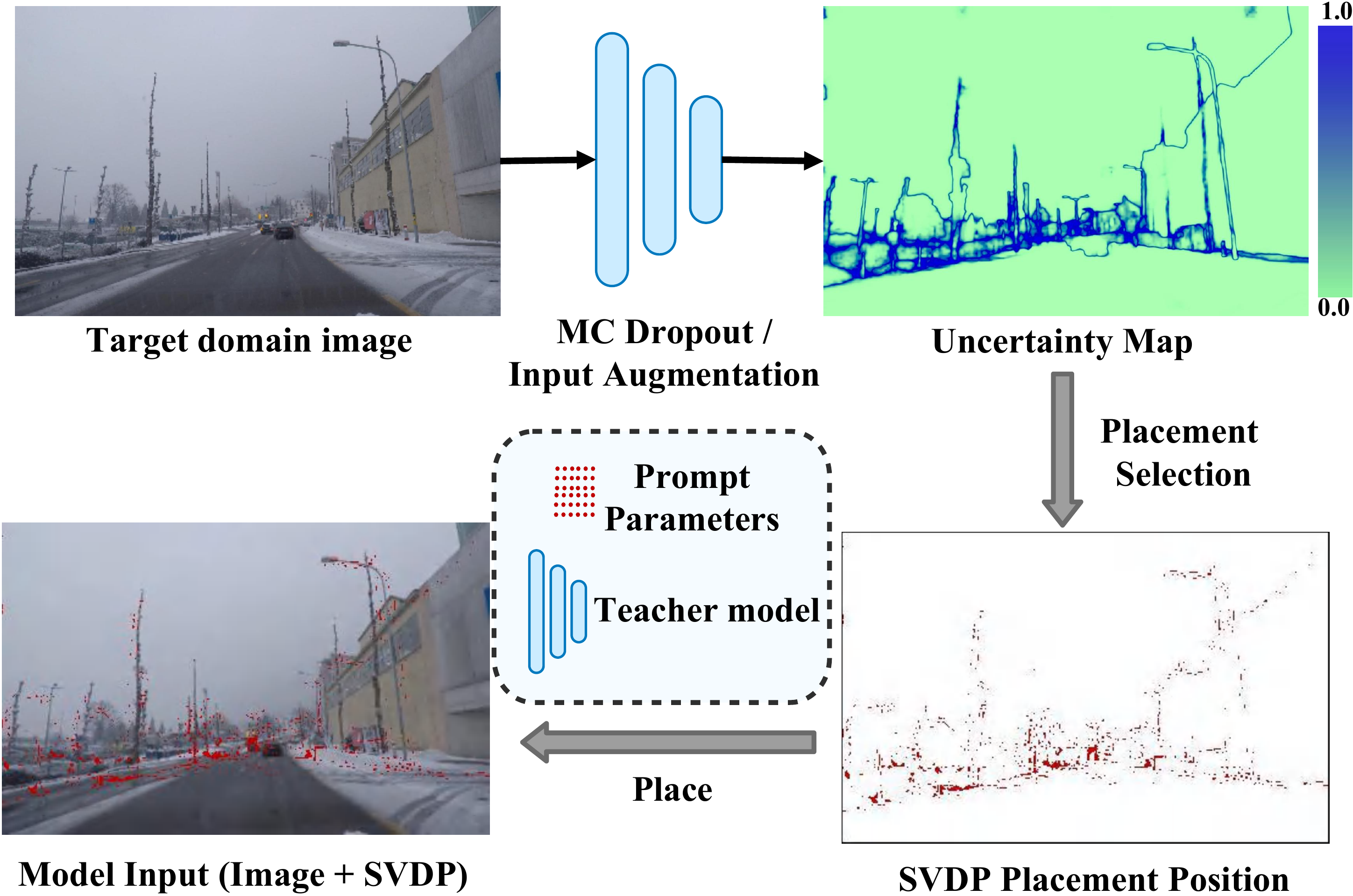}
\centering
% \vspace{-0.15cm}
\caption{The detailed process of Domain Prompt Placing. The uncertainty map is estimated by MC Dropout \cite{gal2016dropout}. The SVDP parameters are placed on the pixels with high uncertainty, then warp into the raw image.}
\label{fig:DPP}
% \vspace{-0.15cm}
\end{figure}

We propose the Domain Prompt Placement (DPP) strategy of SVDP to efficiently extract local domain knowledge in pixel-wise. We intend to place trainable parameters of SVDP on the
pixel with large distribution shifts and adapt pixel-wise data distribution from source to target domain. 
As shown in Fig. \ref{fig:DPP}, we employ the MC Dropout method \cite{gal2016dropout} to perform $m$ forward propagations ($m = 10$) and obtain $m$ group probabilities for each pixel. Specifically, dropout operation is only applied to the linear layer within the prediction head. Calculating the uncertainty value does not significantly increase the computational cost. 
Meanwhile, we can also obtain $m$ sets of probabilities through the simpler method of input image resolution augmentation.
Inspired by \cite{roy2022uncertainty,gan2022cloud}, we calculate the uncertainty value (Eq.\eqref{eq:mc}) of the input and figure out the pixel-wise degree of domain shift.
\begin{equation}
\mathcal{U} (\widetilde{x}_j) =  \left( \frac{1}{m} \sum_{i=1}^m \|p_i(\widetilde {y_j}|\widetilde {x_j}) - \mu \|^2 \right) ^{\frac{1}{2}}
\label{eq:mc}
\end{equation}
, where $p_i(\widetilde{y_j}|\widetilde{x_j})$ is the predicted probability of input pixel $\widetilde{x_j}$ in the $i^{th}$ forward propagation, and $\mu$ is the mean prediction ($m$ rounds) of $\widetilde{x_j}$. $\mathcal{U} (\widetilde{x_j})$ thus represents the uncertainty of the source model for pixel-wise target input $\widetilde{x_j}$. As shown in the bottom of Fig .\ref{fig:DPP}, we sort all pixels based on their pixel-wise uncertainty value and place prompt parameters on the pixels with high uncertainty scores.

\subsection{Domain prompt updating}
\label{sec:3.5}
Motivated by the fact that the mean teacher predictions have a higher quality than the standard model \cite{tarvainen2017mean}, we utilize a mean-teacher model to provide more accurate predictions in the TTA process. To be specific, we adopt the widely-used exponential moving average (EMA) to achieve the model and prompt updating. Same as previous works~\cite{Wangetal2022}, the teacher model ($\mathcal{T}_{mean}$) is updated by EMA from the student model ($\mathcal{S}_{target}$), shown in Eq. \eqref{eq:ema}:
\begin{equation}
 \mathcal{T}_{mean}^{t} = \alpha \mathcal{T}_{mean}^{t-1} + (1-\alpha) \mathcal{S}_{target} ^{t}
\label{eq:ema}
\end{equation}
When $t = 0$ ($t$ is the time step), we utilize the source domain pre-trained model to initialize the weight of the teacher and student model. And we set $\alpha = 0.999$ \cite{AnttiTarvainenetal2017}, which is the updating weight of EMA. 

\begin{figure}[t]
\includegraphics[width=0.45\textwidth]{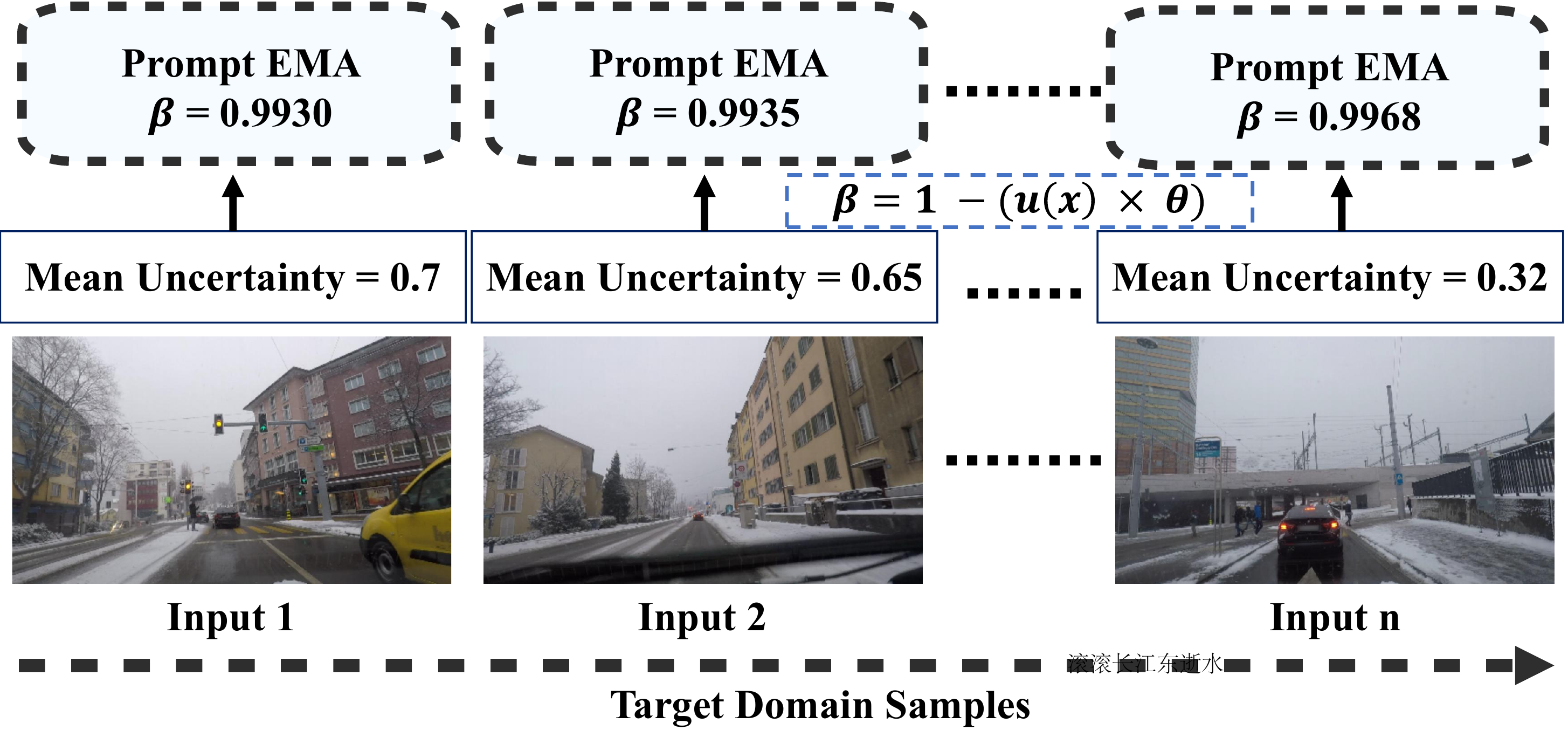}
\centering
% \vspace{-0.25cm}
\caption{The process of Domain Prompt Updating. We adaptively adjust the prompt EMA updating rate for each target domain sample based on image-level uncertainty value.}
\label{fig:DPU}
% \vspace{-0.15cm}
\end{figure}

Different from traditional model updating, we design a special Domain Prompt Updating (DPU) strategy for SVDP to stably fit in target domain distribution. As shown in Fig .\ref{fig:DPU}, we adopt image-level uncertainty value to reflect the degree of domain shift for each target domain sample. 
We calculate the image-level uncertainty value $\mathcal{U}(x)$ by average the pixel-wise uncertainty, shown in Eq. \eqref{eq:unc_for_promptregion}:
\begin{equation}
\mathcal{U}(x) = \frac{1}{H \times W}
\sum_{j }^{H \times W} \mathcal{U} (\widetilde{x}_j)
\label{eq:unc_for_promptregion}
% \vspace{-0.15cm}
\end{equation}
Based on the image-level uncertainty score, we update prompt parameters for each sample with different weight. 

\begin{equation}
\label{unc_ema_prompt}
\begin{aligned}
p_{t} = \beta p_{t-1} + (1-\beta) p_{t},
\end{aligned}
\end{equation}
Note that, $p_{t}$ represents the parameters of the SVDP that is updated by Eq. \eqref{eq:loss}. In DPU, we set the prompt EMA updating rate $\beta = 1-(\mathcal{U}(x) \times \theta)$. $\theta$ is intended to bring the value of uncertainty up to the same order of magnitude as the value of the common EMA update parameter (e.g., $\theta = 0.01$).
As shown in the top of Fig .\ref{fig:DPU}, the prompt EMA weight is set to a large value when the input is of high uncertainty score since the large weight can efficiently adapt to the sample with the large data distribution shift.

\subsection{Loss function}
We utilize teacher model to generate the pseudo labels ($\widetilde{y}_t$), which is refined by test-time augmentation and confidence filter ~\cite{Wangetal2022}.
Then, we adopt consistency loss ($L_{con}$) as the optimization objective for segmentation task, which is a pixel-wise cross-entropy loss \cite{xie2021segformer}.
\begin{equation}
 \mathcal{L}_{con}(\widetilde {x}) = -
 \frac{1}{H \times W} \sum_{w,h}^{W,H}
 \sum_c^C \widetilde{y}_t(w,h,c) \log \hat{y}_t(w,h,c)
\label{eq:loss}
\end{equation}
Where $\hat{y}_t$ is the output of our student model, $C$ means the amount of categories. The loss function of depth estimation is shown in the supplement.

%%%%%%%%%%%%%%%%%%%%%%%%%%%%%%%
\section{Experiments}

\begin{table*}[htb]
\centering
\setlength\tabcolsep{11pt}
\begin{adjustbox}{width=1\linewidth,center=\linewidth}
\begin{tabular}{c|c|cc|cc|cc|cc|c }
\hline

\multicolumn{2}{c|}{Test-Time Adaptation}          & \multicolumn{2}{c|}{Source2Fog}    & \multicolumn{2}{c|}{Source2Night}     & \multicolumn{2}{c|}{Source2Rain}  & \multicolumn{2}{c|}{Source2Snow}    & \multirow{2}{*}{Mean-mIoU}  \\ \cline{1-10}
Method &REF &mIoU$\uparrow$ &mAcc$\uparrow$ 
&mIoU$\uparrow$ &mAcc$\uparrow$ &mIoU$\uparrow$ &mAcc$\uparrow$ &mIoU$\uparrow$ &mAcc$\uparrow$& \\ \hline
Source & NIPS2021 \cite{xie2021segformer}&69.1&79.4&40.3&55.6&59.7&74.4&57.8&69.9 &56.7\\ 
TENT  & ICLR2021 \cite{DequanWangetal2021}  &69.0&79.5&  40.3&55.5&  59.9&74.1&  57.7&69.7 &56.7\\ 
CoTTA& CVPR2022\cite{Wangetal2022} &70.9&80.2 &41.2&55.5 &62.6&75.4 &59.8&70.7&58.6\\ 
DePT & ICLR2023\cite{gao2022visual}  &71.0&80.2&40.9&\textbf{55.8}&61.3&74.4&59.5&70.0&58.2\\ 
VDP & AAAI2023\cite{gan2022decorate}  &70.9&80.3&41.2&55.6&62.3&75.5&59.7&70.7&58.5\\ 

\cellcolor{lightgray}\textbf{SVDP} &\cellcolor{lightgray}\textbf{ours} &\cellcolor{lightgray}\textbf{72.1}&\cellcolor{lightgray}\textbf{81.2} &\cellcolor{lightgray}\textbf{42.0}&\cellcolor{lightgray}54.9& \cellcolor{lightgray}\textbf{64.4}&\cellcolor{lightgray} \textbf{76.7} &\cellcolor{lightgray}\textbf{62.2}&\cellcolor{lightgray}\textbf{72.8}&\cellcolor{lightgray}\textbf{60.1$\pm$0.2}\\ 
 \hline
\end{tabular}
\end{adjustbox}
% \vspace{-0.1cm}
\caption{\textbf{Performance comparison of Cityscapes-to-ACDC TTA.} We use Cityscape as the source domain and ACDC as the four target domains in this setting. 
Mean-mIoU represents the average mIoU value in four TTA experiments.
}
\label{tab:TTA}
% \vspace{-0.1cm}
\end{table*}

\begin{table*}[htb]
\centering
\setlength\tabcolsep{2pt}
\begin{adjustbox}{width=1\linewidth,center=\linewidth}
\begin{tabular}{c|c|ccccc|ccccc|ccccc|c|c }
\hline

\multicolumn{2}{c|}{Time}     & \multicolumn{15}{c}{$t$ \makebox[10cm]{\rightarrowfill} }                                                                              \\ \hline
\multicolumn{2}{c|}{Round}          & \multicolumn{5}{c|}{1}    & \multicolumn{5}{c|}{2}     & \multicolumn{5}{c|}{3}  & \multirow{2}{*}{Mean$\uparrow$}   & \multirow{2}{*}{Gain}  \\ \cline{1-17}
Method & REF & Fog & Night & Rain & Snow & Mean$\uparrow$ & Fog & Night & Rain & Snow  & Mean$\uparrow$ & Fog & Night & Rain & Snow & Mean$\uparrow$ & \\ \hline
Source & NIPS2021 \cite{xie2021segformer}  &69.1&40.3&59.7&57.8&56.7&69.1&40.3&59.7& 	57.8&56.7&69.1&40.3&59.7& 57.8&56.7&56.7&/\\ 
TENT & ICLR2021 \cite{DequanWangetal2021}  &69.0&40.2&60.1&57.3&56.7&68.3&39.0&60.1& 	56.3&55.9&67.5&37.8&59.6&55.0&55.0&55.7&-1.0\\ 
CoTTA & CVPR2022\cite{Wangetal2022}  &70.9&41.2&62.4&59.7&58.6&70.9&41.1&62.6& 	59.7&58.6&70.9&41.0&62.7&59.7&58.6&58.6&+1.9\\ 
DePT & ICLR2023\cite{gao2022visual} 
&71.0&40.8&58.2&56.8&56.5&68.2&40.0&55.4&53.7& 54.3&66.4&38.0&47.3&47.2&49.7&53.4&-3.3\\
VDP & AAAI2023\cite{gan2022decorate}  &70.5&41.1&62.1&59.5&  58.3    &70.4&41.1&62.2&59.4& 58.2     & 70.4&41.0&62.2&59.4& 58.2   &  58.2 & +1.5\\
\cellcolor{lightgray}\textbf{SVDP} &\cellcolor{lightgray}\textbf{ours} &\cellcolor{lightgray}\textbf{72.1}&\cellcolor{lightgray}\textbf{44.0}&\cellcolor{lightgray}\textbf{65.2}&\cellcolor{lightgray}\textbf{63.0}&\cellcolor{lightgray}\textbf{61.1}& 
 \cellcolor{lightgray}\textbf{72.2}&\cellcolor{lightgray}\textbf{44.5}&\cellcolor{lightgray}\textbf{65.9}&\cellcolor{lightgray}\textbf{63.5}&\cellcolor{lightgray}\textbf{61.5} 
 &\cellcolor{lightgray}\textbf{72.1}&\cellcolor{lightgray}\textbf{44.2}&\cellcolor{lightgray}\textbf{65.6}&\cellcolor{lightgray}\textbf{63.6}&\cellcolor{lightgray}\textbf{61.4}      &\cellcolor{lightgray}\textbf{61.1$\pm$0.3} 
 &\cellcolor{lightgray}+\textbf{4.4$\pm$0.3}\\\bottomrule
 \hline
\end{tabular}
\end{adjustbox}
% \vspace{-0.1cm}
\caption{\textbf{Performance comparison for Cityscape-to-ACDC CTTA.} We take the Cityscape as the source domain and ACDC as the continual target domains. During testing, we sequentially evaluate the four target domains three times. Mean is the average score of mIoU. Gain refers to the improvement achieved by the method compared to the Source model.}
% \vspace{-0.1cm}
\label{tab:CTTA}
\end{table*}

In the first subsection, we provide the details of the task settings for test-time adaptation (TTA) and continual TTA (CTTA), as well as a description of the datasets. In the second and third subsections, we compare our method with other baselines \cite{xie2021segformer, DequanWangetal2021, Wangetal2022, gao2022visual, gao2022visual} in four TTA and two CTTA scenarios. 
Comprehensive ablation studies are conducted in the last subsection, which investigates the impact of each component.
We also provide detailed quantitative analysis and supplementary qualitative analysis in the Appendix.

\subsection{Task settings and Datasets}
\label{sec:4.1}
\textbf{TTA and CTTA} are commonly used source-free technology in real-world scenarios in which a source pre-trained model adapts to the distribution of an unseen target domain \cite{liang2023comprehensive}. CTTA is of the same setting as TTA but further sets the target domain constantly changing, bringing more difficulties during the continual adaptation process.

\textbf{Cityscapes-to-ACDC} is designed for semantic segmentation cross-domain learning. And we conduct four TTA and one CTTA experiment on the scenario. The source model is an off-the-shelf pre-trained segmentation model that was trained on the Cityscapes dataset~\cite{cordts2016cityscapes}. 
The ACDC dataset~\cite{sakaridis2021acdc} contains images collected in four different unseen visual conditions: Fog, Night, Rain, and Snow. For the TTA, we adapt the source pre-trained model to each of the four ACDC target domains separately. For the CTTA, we repeat the same sequence of target domains (Fog→Night→Rain→Snow) multiple times to simulate environment changes in real-world scenarios \cite{Wangetal2022}.

\textbf{KITTI-to-Driving Stereo.}
To demonstrate the generalization of our method, we also conduct experiments in depth estimation CTTA scenario. The source model employed is an off-the-shelf, pre-trained model, initially trained on the KITTI dataset \cite{geiger2012we}. The Driving Stereo\cite{yang2019drivingstereo} comprises images collected under four disparate, unseen visual conditions: foggy, rainy, sunny, and cloudy. For the CTTA, we repeat the same sequence of target domains (Foggy→Rainy→Sunny→Cloudy) multiple times.

\textbf{Implementation Details.} We follow the implementation details \cite{Wangetal2022} to set up our semantic segmentation TTA experiments. Specifically, we use the Segformer-B5 ~\cite{xie2021segformer} pre-trained on Cityscapes as our off-the-shelf source model.  
We down-sample the original image size of 1920x1080 of the ACDC dataset to 960x540, which serves as network input. We evaluate our predictions under the original resolution. 
We use a range of image resolution scale factors [0.5, 0.75, 1.0, 1.25, 1.5, 1.75, 2.0] for the augmentation method in the teacher model. 
For depth estimation CTTA, we follow the implementation details in previous work~\cite{liu2022unsupervised} and adopt the pre-trained DPT~\cite{Ranftl2021} on the KITTI as the source model. 
The optimizer is performed using Adam optimizer~\cite{kingma2014adam}  with $(\beta_1, \beta_2) = (0.9, 0.999)$. 
We set the batch size 1 for both TTA and CTTA experiments and all experiments are conducted on NVIDIA A100 and 3090 GPUs.
\subsection{The effectiveness on Semantic Segmentation}
\label{sec:4.2}

\textbf{Cityscapes-to-ACDC TTA.}
We evaluate the performance of the proposed SVDP on four scenarios with significant domain gap during TTA. 
Tab .\ref{tab:TTA} shows that the Mean-mIoU for the four domains using the source domain model alone is only 56.7\%. Recent advanced methods CoTTA increases it to 58.6\% while our method further increases it by 1.5\%. 
These results demonstrate that our method can better address the domain shit problem in test time compared to other methods.  Furthermore, in contrast to VDP, which employs dense prompts, our method successfully circumvents the occlusion issue, leading to improved extraction of both semantic and domain knowledge for TTA. In comparison to DePT, which introduces prompts at the token level, our SVDP approach operates at the image-level. This aspect enables the extraction of local domain knowledge, thereby resulting in substantial performance enhancements.

\begin{figure*}[htb]
\centering
\includegraphics[width=0.98\linewidth]{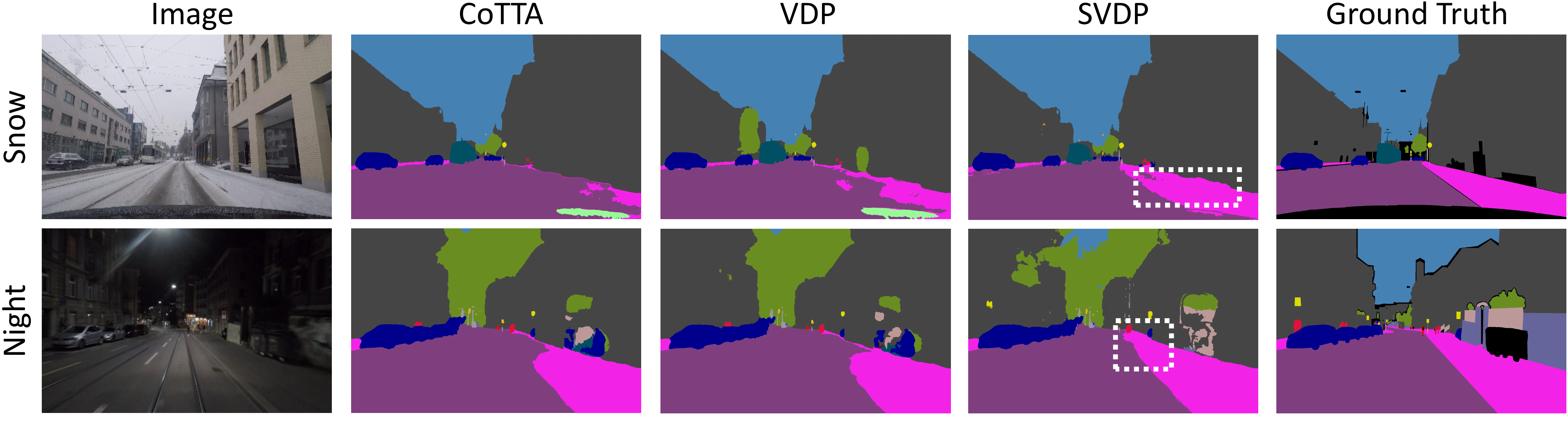}
% \vspace{-0.25cm}
\caption{Qualitative comparison of our method with previous SOTA methods on the ACDC dataset. Our method could better segment different pixel-wise classes such as shown in the white box. 
}
\label{fig:vis}
% \vspace{-0.1cm}
\end{figure*}

\textbf{Cityscapes-to-ACDC CTTA.}
To demonstrate that our method can also address continuously changing domain shifts, we deal with the four domain data during test time periodically.
As shown in Tab .\ref{tab:CTTA}, due to catastrophic forgetting, the performance of TENT and DePT gradually decreases over time. These methods only focus on acquiring new domain-specific knowledge from the target domain, resulting in a neglect of the original knowledge from the source domain.
And we find that our method gains 2.5\% increase of mIoU more than the previous SOTA CTTA method \cite{Wangetal2022}. 
The results prove that our method can continuously extract target domain knowledge via sparse prompt and preserve previous domain knowledge via model parameters, showing the ability to address dynamic domain shifts. 
In term of qualitative analysis, shown in Fig .\ref{fig:vis}, our method correctly distinguishes the sidewalk from the road, avoiding mis-classification in target domains. 

Overall, our method outperforms several previous SOTA methods on all semantic segmentation TTA and CTTA tasks and shows promising potential for real-world applications.

\begin{table*}[htb]
\centering
\setlength\tabcolsep{2pt}%调列距
\begin{adjustbox}{width=1\linewidth,center}
\begin{tabular}{c|cccc|cccc|cccc|cccc|c }
\hline

\multicolumn{1}{c|}{Time}     & \multicolumn{14}{c}{$t$ \makebox[13.8cm]{\rightarrowfill} }                                                                              \\ \hline
\multicolumn{1}{c|}{Domain}          & \multicolumn{4}{c|}{Foggy}    & \multicolumn{4}{c|}{Rainy}     & \multicolumn{4}{c|}{Sunny} & \multicolumn{4}{c|}{Cloudy} & \multirow{2}{*}{Mean$\uparrow$}     \\ \cline{1-17}
Method  & $\delta>1.25$$\uparrow$ & $\delta>1.25^2$$\uparrow$ & AbsRel$\downarrow$ & RMSE$\downarrow$ &  $\delta>1.25$ $\uparrow$& $\delta>1.25^2$$\uparrow$ & AbsRel$\downarrow$ & RMSE$\downarrow$ &  $\delta>1.25$$\uparrow$ & $\delta>1.25^2$ $\uparrow$& AbsRel$\downarrow$ & RMSE$\downarrow$ &  $\delta>1.25$$\uparrow$ & $\delta>1.25^2$$\uparrow$ & AbsRel$\downarrow$ & RMSE$\downarrow$ &  \\ \hline
Source   &0.040&0.791&0.313&10.864&0.046&0.573&0.382&15.966& 0.153&0.896&0.270&7.404&0.134& 0.8697&0.282&7.758&0.093\\ 
CoTTA   &0.436&0.892&0.268&9.615&0.553&0.791&0.289& 12.231&0.597&0.964&0.191&5.445&0.476&0.959&0.203&5.536&0.516\\ 
VDP  &0.683&0.902&0.159&9.068& 0.575&0.692&0.253&15.403& 0.716&0.876&0.181&8.905&0.689&0.858&0.197& 9.200&0.666\\
\cellcolor{lightgray}\textbf{SVDP} &\cellcolor{lightgray}\textbf{0.734}&\cellcolor{lightgray}\textbf{0.943}&\cellcolor{lightgray}\textbf{0.152}&\cellcolor{lightgray}\textbf{7.092}&\cellcolor{lightgray}\textbf{0.689}&\cellcolor{lightgray}\textbf{0.894}&\cellcolor{lightgray}\textbf{0.159}&\cellcolor{lightgray}\textbf{9.700}&\cellcolor{lightgray}\textbf{0.800}&\cellcolor{lightgray}\textbf{0.937}&\cellcolor{lightgray}\textbf{0.168}&\cellcolor{lightgray}\textbf{5.719}&\cellcolor{lightgray}\textbf{0.784}&\cellcolor{lightgray}\textbf{0.930}&\cellcolor{lightgray}\textbf{0.169}&\cellcolor{lightgray}\textbf{5.784}&\cellcolor{lightgray}\textbf{0.741$\pm$0.03}\\\bottomrule
\hline
\end{tabular}
\end{adjustbox}
% \vspace{-0.3cm}
\caption{\textbf{Performance comparison for KITTI-to-Driving Stereo CTTA.} We take the KITTI as the source domain and Driving Stereo as the continual target domains. Mean is the average score of $\delta>1.25$ for four domains.}
% \vspace{-0.1cm}
\label{tab: Driving Stereo}
\end{table*}

\subsection{The effectiveness on Depth Estimation}
\textbf{KITTI-to-Driving Stereo CTTA.}
To demonstrate the effectiveness of our approach in addressing CTTA problem in depth estimation task, we conducted a series of evaluations on four distinct target domains from the Driving Stereo dataset at regular intervals during the testing phase.
As shown in Table \ref{tab: Driving Stereo}, our method consistently outperforms the state-of-the-art (SOTA) technique across all four evaluation metrics. Particularly noteworthy is the significant enhancement in the mean $\delta>1.25$, achieving a remarkable improvement of 65.9\% when compared to the Source model, and an impressive 7.5\% improvement over the previous SOTA method. This result underscores the robust continual adaptation ability of our method in the context of depth estimation.
Given that CTTA has access to the data only once, as opposed to CoTTA, our approach leverages sparse prompt to effectively adapt to the target domain, resulting in significant performance gains.  
Overall, these results show that our SVDP consistently attains superior outcomes in the depth estimation tasks.
\subsection{Ablation study}
\label{sec:4.3}
In this subsection, we evaluate the contribution of each component in our method. Since the CTTA is the most challenging and realistic scenario, we conduct the ablation study on the KITTI-to-Driving Stereo CTTA. 
% Due to the space limitations, more ablation study is shown in the Appendix.
\begin{table}[!tb]
\label{ablationDAP}
\centering
\setlength\tabcolsep{4.0pt}%调列距
\renewcommand\arraystretch{1}%调行距
\begin{tabular}{c|cccc|cc}
\toprule
 & \makecell*[c]{TS} & \makecell*[c]{SVDP} & \makecell*[c]{DPP} & \makecell*[c]{DPU}  & Abs Rel$\downarrow$ & $\delta>1.25 
 \uparrow$\\\midrule
$Ex_{1}$ &  & & & &0.312& 0.093 \\ 
$Ex_{2}$& \checkmark &  & &  & 0.249& 0.503\\
$Ex_{3}$ &\checkmark  & \checkmark &  & & 0.187& 0.705\\
$Ex_{4}$  & \checkmark & \checkmark &\checkmark  &  &0.169& 0.737\\
$Ex_{5}$  & \checkmark & \checkmark & &\checkmark & 0.177& 0.723\\
$Ex_{6}$ & \checkmark &  \checkmark &\checkmark  &\checkmark & 0.162& 0.741\\
\bottomrule
\end{tabular}
% \vspace{-0.2cm}
\caption{\textbf{Ablation: Contribution of each component. % need to be modified lower is better
}}
% \vspace{-0.3cm}
\label{tab:ablation}
\end{table}
\textbf{Effectiveness of each component.} 
As presented in Tab.~\ref{tab:ablation} $Ex_{2}$, 
Teacher-student~(TS) structure is a common technique in CTTA \cite{Wangetal2022, gan2022decorate}, which is used to generate pseudo labels in the target domain and only has 0.063 Abs Rel reduces without our method. This verifies the improvement of our method does not come from the usage of this prevalent scheme.
In $Ex_{3}$, by introducing sparse prompts (SVDP), we observe that the Abs Rel reduces 0.062 and $\delta>1.25$ increases 20.2\%, respectively. The result demonstrates that SVDP facilitates addressing the domain shift problem, since it can extract local target domain knowledge without damaging the original semantic information.
As shown in $Ex_{4}$, DPP achieves further 0.018 Abs Rel reduces and 3.2\% $\delta>1.25$ improvement since the specially designed prompt placement strategy can assist SVDP in extracting target domain-specific knowledge more efficiently.  
Compared with $Ex_{3}$, DPU ($Ex_{5}$) also reduces the Abs Rel 0.01 and improves 1.8\% $\delta>1.25$, respectively. The results prove the effectiveness of DPU and show the importance of adaptively optimizing for different samples during TTA process.
$Ex_{6}$ shows the complete combination of  all components which achieves 64.8\% $\delta>1.25$ improvement and 0.150 Abs Rel reduction in total. It proves that all components compensate each other and jointly address the depth estimation domain shift problem in test time.

\begin{figure}[t]
\includegraphics[width=0.4\textwidth]{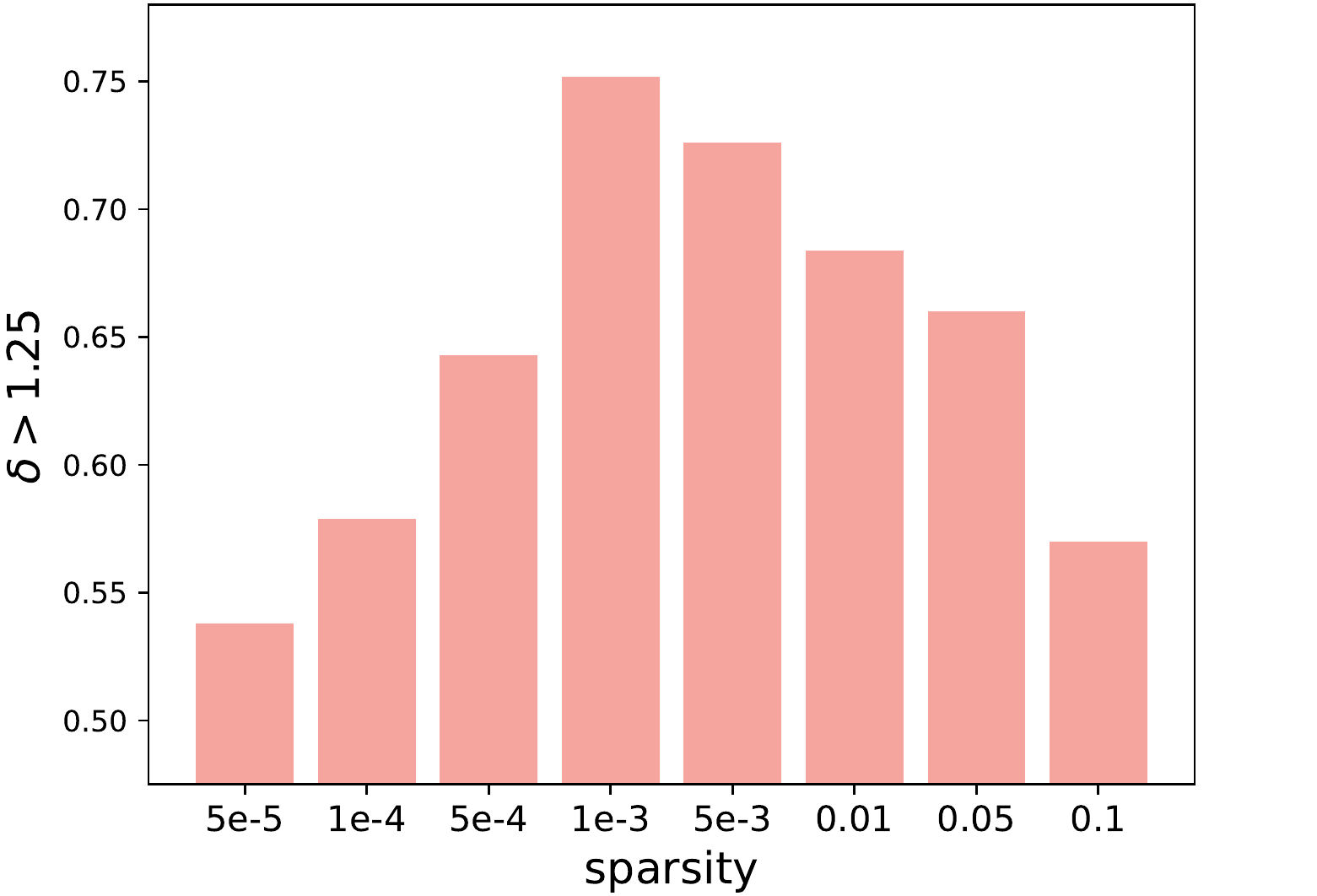}
\centering
% \vspace{-0.6cm}
\caption{Effect of prompts' sparsity}
\label{fig:exp_prompt_sparsity}
% \vspace{-0.2cm}
\end{figure}

\textbf{How does the prompt sparsity affect the performance?}
As shown in Fig .\ref{fig:exp_prompt_sparsity}, we investigate the performance impact caused by the sparsity of SVDP.  Specifically, we gradually increase the density of SVDP pixel-wise parameters and place it into more pixels. We find that $\delta>1.25$ initially improves along with increasing SVDP density and then starts to decrease when the density exceeds 1e-3.
This observation suggests that when SVDP is excessively sparse, it fails to capture the domain-specific knowledge effectively due to the limited number of parameters. In contrast, if the SVDP becomes too dense, the prompt will occlude many spatial details, leading to depth estimation performance degradation.
Therefore, it is crucial to strike a balance on the degree of prompt sparsity and we consider that SVDP can achieve optimal potential in 1e-3 sparsity.

\section{Conclusion}
In this paper, we are the first to introduce the Sparse Visual Domain Prompt (SVDP) in dense prediction TTA tasks (i.e., semantic segmentation, depth estimation), which address the problem of inaccurate contextual information extraction and insufficient domain-specific feature transferring caused by dense prompt occlusion. Moreover, the Domain Prompt Placement (DPP) and Domain Prompt Updating (DPU) strategies are specially designed for applying SVDP to ease the pixel-wise domain shift better. Our method demonstrates state-of-the-art performance and effectively addresses domain shift through extensive experimentation across various TTA and CTTA scenarios.

\paragraph{Acknowledgments.}
Shanghang Zhang is supported by the National Key Research and Development Project of
China (No.2022ZD0117801). This work was supported by the Emerging Engineering Interdisciplinary Youth Program at Peking University under Grant 7100604372

% \clearpage

\bibliography{aaai24}

\clearpage
\appendix

%%%%%%%%% ABSTRACT
\begin{figure*}[ht!]
\centering
\includegraphics[width=\linewidth]{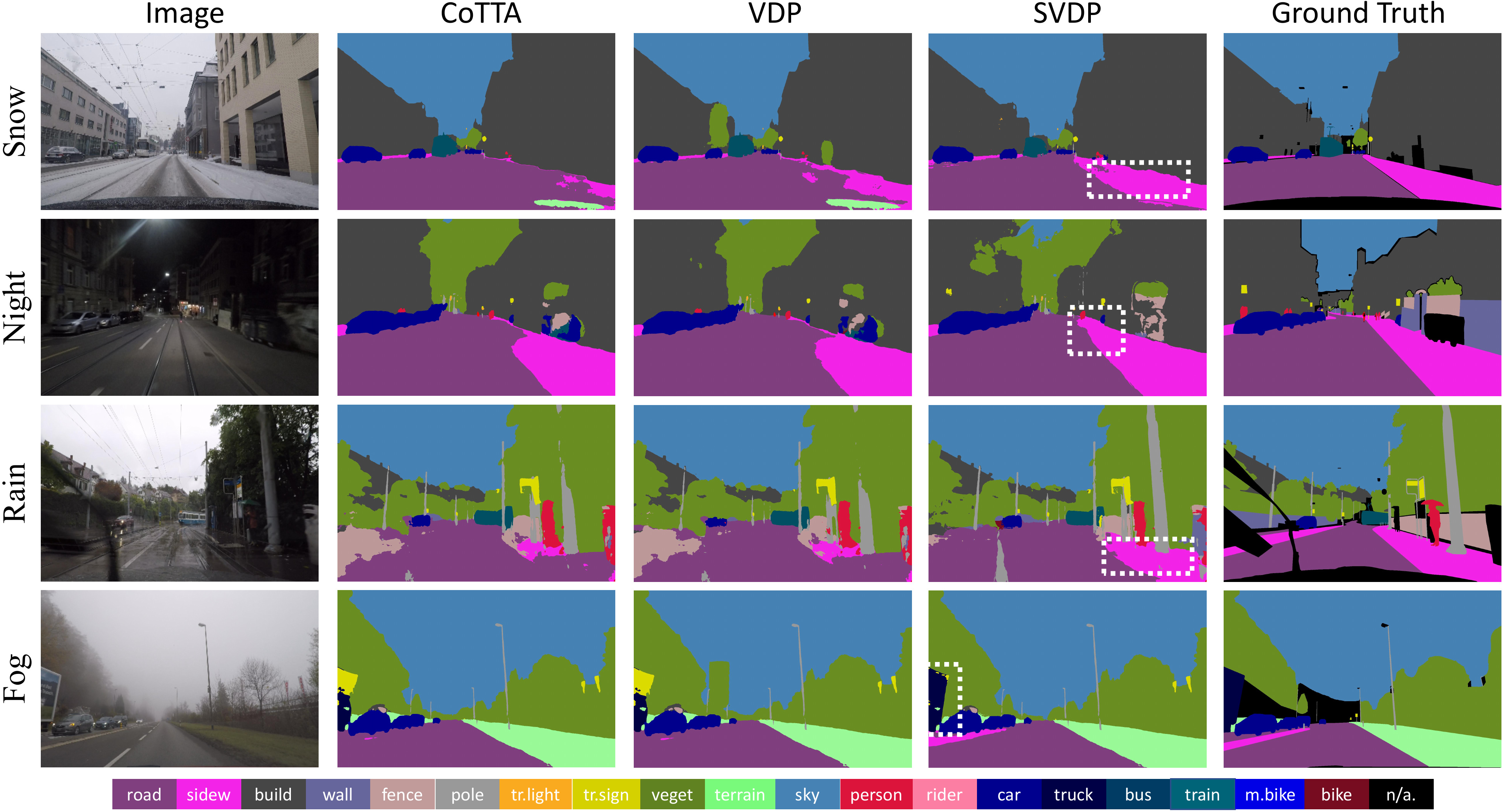}
\vspace{-0.39cm}
\caption{Qualitative comparison of SVDP with previous SOTA method: CoTTA~\cite{Wangetal2022}, VDP~\cite{gan2022decorate} on ACDC Fog, Night, Rain, and Snow four scenarios. SVDP could better segment different pixel-wise classes such as shown in the white box.}
\label{fig:qualitative} 
\vspace{-0.3cm}
\end{figure*}
\section*{Appendix Overview}
\label{sec: ap}

The following items are included in this supplementary material.

\begin{itemize}
    \item Additional Ablation Studies of
    \begin{itemize}
        \item Depth Estimation on TTA scenario
        \item Domain Prompt Updating
    \end{itemize}
    \item Qualitative Analysis of 
    \begin{itemize}
        \item Semantic Segmentation
        \item Depth Estimation
    \end{itemize}
    \item Additional Quantitative Results
    \item Additional Related Work
\end{itemize}

\section{Additional Ablation Studies}
\subsection{Depth Estimation on TTA scenario}
\label{Sec:ablation ctta}
The proposed method comprises a Sparse Visual Prompt (SVDP), a Domain Prompt Placement (DPP) strategy, and a Domain Prompt Updating (DPU) strategy to mitigate domain shifts in semantic segmentation tasks. In this study, we conduct ablation experiments on the TTA scenario (KITTI-to-DrivingStereo Foggy) to evaluate the effectiveness of each component.

To compare the performance of our method with and without using the teacher-student (TS) structure, a common technique in TTA used to generate pseudo labels in the target domain, we present the results in Tab.~\ref{aptab:ablation} $Ex_{2}$. The results show that without our method, TS only has 0.069 Abs Rel reduces, indicating that our method's improvement does not come from the usage of this prevalent scheme.
In $Ex_{3}$, we introduce SVDP to extract local target domain knowledge without damaging the original spatial information. The results demonstrate that SVDP achieves a 21.0\% $\delta>1.25$ improvement and 0.073 Abs Rel reduction, effectively addressing the domain shift problem. In $Ex_{4}$, DPP achieves a further 3.3\% $\delta>1.25$ improvement and 0.014 Abs Rel reduces by serving as a specially designed prompt placement strategy to assist SVDP in extracting more target domain-specific knowledge.
We evaluate the effectiveness of the DPU in $Ex_{5}$, which adaptively optimizes for different samples during testing. Compared with $Ex_{3}$, DPU reduces the Abs Rel 0.006 and improves 2.2\% $\delta>1.25$ respectively. Finally, in $Ex_{6}$, we show the complete combination of all components, which achieves a total of 69.4\% $\delta>1.25$ improvement and 0.161 Abs Rel reduction. These results demonstrate that all components of our method effectively address the depth estimation domain shift and compensate for each other to achieve superior performance.

\begin{table}[!tb]
\caption{\textbf{Ablation: Contribution of each component on KITTI-to-DrivingStereo Foggy. 
}}
\centering
\setlength\tabcolsep{5pt}%调列距
\renewcommand\arraystretch{1}%调行距
\begin{tabular}{c|cccc|cc}
\toprule
 & \makecell*[c]{TS} & \makecell*[c]{SVDP} & \makecell*[c]{DPP} & \makecell*[c]{DPU}  & Abs Rel$\downarrow$ & $\delta>1.25 \uparrow$\\\midrule
$Ex_{1} $ &  & & & &0.313& 0.040 \\ 
$Ex_{2} $& \checkmark &  & &  & 0.244& 0.482\\
$Ex_{3}$ &\checkmark  & \checkmark &  & & 0.171& 0.692\\
$Ex_{4}$  & \checkmark & \checkmark &\checkmark  &  &0.157& 0.725\\
$Ex_{5}$  & \checkmark & \checkmark & &\checkmark & 0.165& 0.714\\
$Ex_{6} $ & \checkmark &  \checkmark &\checkmark  &\checkmark & 0.152& 0.734\\
\bottomrule
\end{tabular}
\vspace{-0.2cm}
\label{aptab:ablation}
\end{table}

\begin{figure}[!h]
\includegraphics[width=0.43\textwidth]{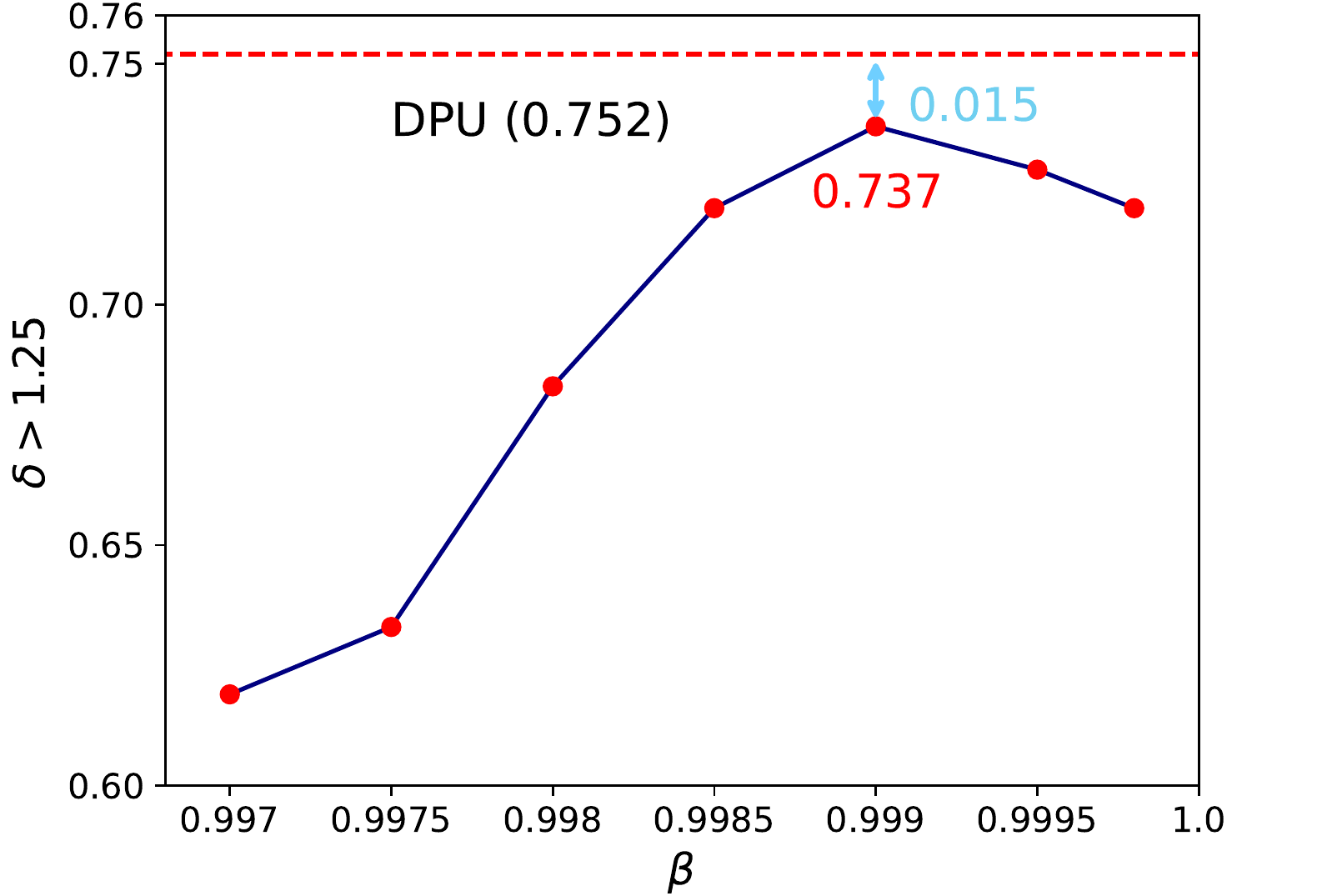}
\centering
\vspace{-0.05cm}
\caption{Sensitivity Analysis: The effect of prompt EMA's parameter $\beta$ on depth estimation performance in the CTTA scenario.}
\label{fig:senstivity}
\vspace{-0.25cm}
\end{figure}
\subsection{Domain Prompt Updating}
\label{Sec: sentivity}
For SVDP, we utilize Eq.5 to update the prompt parameters and conduct an analysis of the sensitivity of the parameter $\beta$ in the depth estimation CTTA scenario. As depicted in Fig.~\ref{fig:senstivity}, we investigate the impact of $\delta>1.25$ values on the performance. Specifically, we gradually increase the value of $\beta$ and record the corresponding $\delta>1.25$ values. We observe that the $\delta>1.25$ improves with increasing $\beta$; however, it starts to decrease once $\beta$ exceeds 0.999. Compare with the best fixed $\beta$ value, our proposed DPU (\textcolor{red}{red line}) strategy can further achieve 1.5\% $\delta>1.25$ improvement. Therefore, due to the different degrees of domain shift, we need to update prompt parameters for the each sample with different EMA weights.

\section{Qualitative analysis}
\subsection{Semantic Segmentation}
\label{Sec:qualitative}
To further demonstrate the effectiveness of our proposed method, SVDP, we conduct a qualitative comparison with two current leading methods, CoTTA \cite{Wangetal2022} and VDP \cite{gan2022decorate}, on the CTTA scenario (Cityscapes-to-ACDC).

The results of the comparison are presented in Fig .\ref{fig:qualitative}. 
In the foggy target-domain, we highlight a white box that contains a tall truck object. This object is difficult to segment as it shares characteristics with the \emph{sign} class. Thanks to the contribution of the Domain Prompt Placement (DPP), our proposed method, SVDP, has a significant advantage in dealing with such confusing semantic segmentation categories with high uncertainty. Our method also outperforms CoTTA and VDP in the remaining three domains. In these domains, our proposed method correctly distinguishes the sidewalk from the road, avoiding misclassification.
Overall, our method can achieve better local segmentation results and neglect the influence of local domain shift.
And our method produces finer results than the previous state-of-the-art methods, with clear visual improvement.

\subsection{Depth estimation}
\begin{figure*}[ht!]
\centering
\includegraphics[width=\linewidth]{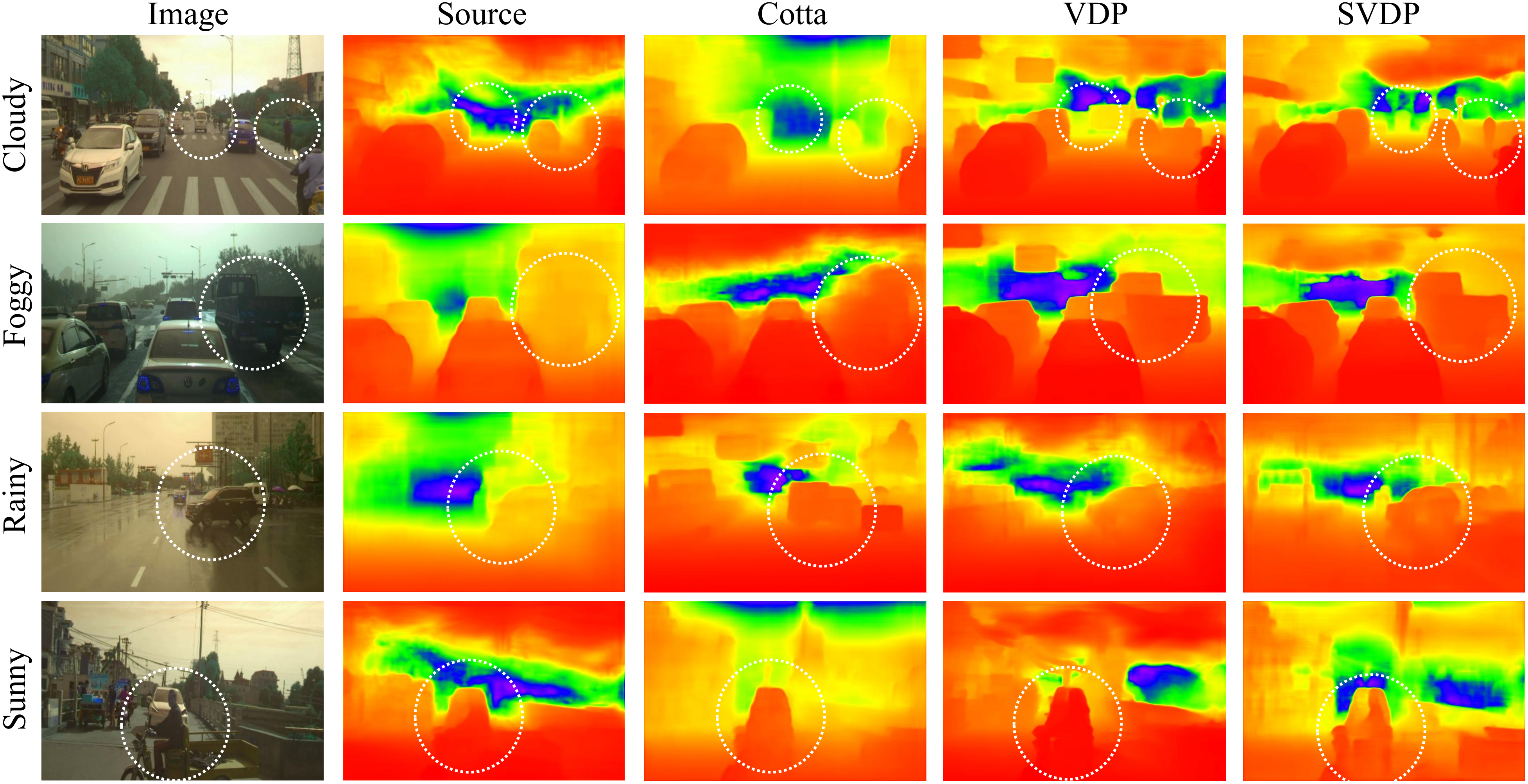}
\vspace{-0.39cm}
\caption{Qualitative comparison of SVDP with previous SOTA method: CoTTA\cite{Wangetal2022}, VDP\cite{gan2022decorate} on Driving Stero  Cloudy, Foggy, Rainy, and Sunny four scenarios. SVDP could make the depth estimation better such as shown in the white circle.}
\label{fig:qualitative_depth} 
\vspace{-0.3cm}
\end{figure*}

To demonstrate the generalizability of our proposed method SVDP, we compare it with the current SOTA method on the Depth Estimation task, and the results of the qualitative analysis are shown in Fig.~\ref{fig:qualitative_depth}

Compared with the VDP and CoTTA, We notice that our SVDP can obviously improves the semantic representation of the model, which concentrates more on the foreground object. 
The results verify that our methods can extract the target domain knowledge during the test time adaptation. 
Furthermore, our SVDP enhances the model's depth estimation ability in the regions of edges, sharpness, and long distances.

\begin{table*}
\caption{Performance Comparison for \textbf{Cityscapes-to-ACDC Fog domain in TTA scenario}. The IoU score of each class and the mIoU score are reported. The best results are highlighted in \textbf{bold}.
  } \label{table:fog acdc}
  \centering
  \resizebox{0.99\textwidth}{!}{
    \def\arraystretch{1.1}
    \begin{tabular}{ l | c c c c c c c c c c c c c c c c c c c | c }
        \Xhline{1.2pt}
        Method & \rotatebox{0}{road} & \rotatebox{0}{side.} & \rotatebox{0}{buil.} & \rotatebox{0}{wall} & \rotatebox{0}{fence} & \rotatebox{0}{pole} & \rotatebox{0}{light} & \rotatebox{0}{sign} & \rotatebox{0}{veg.} & \rotatebox{0}{terr.} & \rotatebox{0}{sky} & \rotatebox{0}{pers.} & \rotatebox{0}{rider} & \rotatebox{0}{car}& \rotatebox{0}{truck} & \rotatebox{0}{bus} & \rotatebox{0}{train} & \rotatebox{0}{mbike} & \rotatebox{0}{bike} & mIoU \\
        \hline
        \hline
        Source  & 94.0 & 63.9 & 79.8 & 55.7 & 24.9 & 45.0 & 41.5 & 69.8 & 86.6 & 71.0 & 97.6 & 64.1 & 66.2 & 87.4 & 73.0 & 92.6 & 87.7 & 50.2 & 61.7 & 69.1 \\
        TENT & 94.0 & 64.0 & 79.7 & 55.4 & 24.6 & 44.6 & 41.4 & 69.9 & 86.7 & 71.1 & 97.6 & 64.0 & 65.9 & 87.4 & 73.0 & 92.6 & 88.0 & 50.2 & 61.9 & 69.0 \\
        CoTTA & 93.9 & 63.6 & 80.0 & 55.5 & 25.1 & 49.0 & 43.4 & 73.0 & 87.0 & 70.7 & 97.8 & 68.6 & 71.3 & 87.1 & 74.8 & 93.6 & 89.1& 58.0 & 66.7 & 70.9 \\
        DePT & 94.0 & 64.0 & 79.9 & 56.1 & 25.3 & 48.8 & 43.5 & 73.0 & 87.1 & 70.6 & 97.5 & 67.9 & \textbf{71.5} & 87.3 & 75.1 & 93.5 & 89.1& 57.4 & 66.5 & 71.0 \\
        VDP & 93.9 & 63.6& 80.0 & 55.6 & 25.1& 49.0& 43.4 & 73.0& 86.9 & 70.7 & 97.7 & 68.5 & 71.1 & 87.2 & 74.7 & 93.5 & 89.2 & 57.9 & 66.6 & 70.9 \\
        \cellcolor{lightgray}\textbf{SVDP} &\cellcolor{lightgray}\textbf{94.4} &\cellcolor{lightgray}\textbf{65.9}&\cellcolor{lightgray}\textbf{80.5}&\cellcolor{lightgray}\textbf{57.8}&\cellcolor{lightgray}\textbf{26.5}&\cellcolor{lightgray}\textbf{50.5}&\cellcolor{lightgray}\textbf{43.9}&\cellcolor{lightgray}\textbf{73.6}&\cellcolor{lightgray}\textbf{87.6}&\cellcolor{lightgray}\textbf{72.0}&\cellcolor{lightgray}\textbf{98.0}&\cellcolor{lightgray}\textbf{68.8}&\cellcolor{lightgray}71.3 
 &\cellcolor{lightgray}\textbf{87.7}&\cellcolor{lightgray}\textbf{77.6}&\cellcolor{lightgray}\textbf{94.4}&\cellcolor{lightgray}\textbf{92.6}&\cellcolor{lightgray}\textbf{60.0}      &\cellcolor{lightgray}\textbf{67.4} 
 &\cellcolor{lightgray}\textbf{72.1}\\\bottomrule

        \Xhline{1.2pt}
    \end{tabular}
  }

\end{table*}

\begin{table*}
\caption{Performance Comparison for \textbf{Cityscapes-to-ACDC Night domain in TTA scenario}. The IoU score of each class and the mIoU score are reported. The best results are highlighted in \textbf{bold}.
  } 
  \centering
  \resizebox{0.99\textwidth}{!}{
    \def\arraystretch{1.1}
    \begin{tabular}{ l | c c c c c c c c c c c c c c c c c c c | c }
        \Xhline{1.2pt}
        Method & \rotatebox{0}{road} & \rotatebox{0}{side.} & \rotatebox{0}{buil.} & \rotatebox{0}{wall} & \rotatebox{0}{fence} & \rotatebox{0}{pole} & \rotatebox{0}{light} & \rotatebox{0}{sign} & \rotatebox{0}{veg.} & \rotatebox{0}{terr.} & \rotatebox{0}{sky} & \rotatebox{0}{pers.} & \rotatebox{0}{rider} & \rotatebox{0}{car}& \rotatebox{0}{truck} & \rotatebox{0}{bus} & \rotatebox{0}{train} & \rotatebox{0}{mbike} & \rotatebox{0}{bike} & mIoU \\
        \hline
        \hline
        Source  & 87.6 & 46.3 & 61.8 & 27.0 & 25.3 & 40.8 & 38.7 & 39.4 & 47.7& 26.8 & \textbf{11.4} & 48.6 & 39.9 & 76.1 & 15.9 & 24.2 & 52.0 & 26.5 & 29.6 & 40.3 \\
        TENT & \textbf{87.7} & 46.4 & 61.9 & 27.1 & 25.2 & 40.8 & 38.8 & 39.3  & 47.0 & 26.8 & 9.6 & 48.7 & 40.0 & 76.2 & 16.1 & 24.3 & 51.9 & 26.6& 29.7 & 40.2 \\
        CoTTA & 87.6 & 46.7 & \textbf{62.3}& 27.2 & 25.0 & 44.0 & 42.9 & 40.8 & 47.2 & 26.7 & 8.8 & 51.8 & 41.9 & 76.6 & 18.8 & 22.4 & 51.7& 27.8 & 32.1 & 41.2 \\
        DePT & 87.3 & 46.5 & 62.0 & 27.0 & 25.3 & 43.5 & 40.9 & 41.0 & 47.2 & 26.6 & 8.8 & 51.0 & 42.5 & 77.1 & 17.5 & 23.0 & 51.5 & 26.4 & 31.7 & 40.9 \\
        VDP & 87.6 & 46.8 & 62.2 & 27.1 & 25.0 & 44.0 & 42.9 & 41.0 & 47.3 & 26.6 & 9.0 & 51.7 & 41.9 & 76.6& 18.7 & 23.2 & 51.9 & 27.8& 32.0 & 41.2 \\
        \cellcolor{lightgray}\textbf{SVDP} &\cellcolor{lightgray}87.5 &\cellcolor{lightgray}\textbf{46.6}&\cellcolor{lightgray}52.7&\cellcolor{lightgray}\textbf{28.3}&\cellcolor{lightgray}\textbf{23.1}&\cellcolor{lightgray}\textbf{46.0}&\cellcolor{lightgray}\textbf{44.7}&\cellcolor{lightgray}\textbf{41.2}&\cellcolor{lightgray}\textbf{56.1}&\cellcolor{lightgray}\textbf{23.9}&\cellcolor{lightgray}10.5&\cellcolor{lightgray}\textbf{53.4}&\cellcolor{lightgray}\textbf{43.2} 
 &\cellcolor{lightgray}\textbf{78.0}&\cellcolor{lightgray}\textbf{25.7}&\cellcolor{lightgray}\textbf{26.0}&\cellcolor{lightgray}\textbf{46.9}&\cellcolor{lightgray}\textbf{29.7}     &\cellcolor{lightgray}\textbf{34.1} 
 &\cellcolor{lightgray}\textbf{42.0}\\\bottomrule

        \Xhline{1.2pt}
    \end{tabular}
  }
\label{table:night acdc}
\end{table*}

\begin{table*}[!htb]
\caption{Performance Comparison for \textbf{Cityscapes-to-ACDC Rain domain in TTA scenario}. The IoU score of each class and the mIoU score are reported. The best results are highlighted in \textbf{bold}.
  } 
  \centering
  \resizebox{0.99\textwidth}{!}{
    \def\arraystretch{1.1}
    \begin{tabular}{ l | c c c c c c c c c c c c c c c c c c c | c }
        \Xhline{1.2pt}
        Method & \rotatebox{0}{road} & \rotatebox{0}{side.} & \rotatebox{0}{buil.} & \rotatebox{0}{wall} & \rotatebox{0}{fence} & \rotatebox{0}{pole} & \rotatebox{0}{light} & \rotatebox{0}{sign} & \rotatebox{0}{veg.} & \rotatebox{0}{terr.} & \rotatebox{0}{sky} & \rotatebox{0}{pers.} & \rotatebox{0}{rider} & \rotatebox{0}{car}& \rotatebox{0}{truck} & \rotatebox{0}{bus} & \rotatebox{0}{train} & \rotatebox{0}{mbike} & \rotatebox{0}{bike} & mIoU \\
        \hline
        \hline
        Source  & 82.3 & 47.1 & 89.5 & 36.8 & 26.6 & 51.0 & 64.8 & 62.9 & 89.5 & 60.3 & 97.8 & 46.0 & 53.0 & 81.1 & 25.3 & 65.4 & 56.7 & 47.6 & 51.2 & 59.7 \\
        TENT & 82.4 & 46.7 & 89.6 & 37.3 & 27.0 & 50.6 & 64.6 & 62.9 & 89.5 & 60.4 & 97.7 & 46.7 &54.5 & 81.2 & 25.4 & 65.2 & 56.6 & 47.3 & 51.8 & 59.9 \\
        CoTTA & 83.0 & 48.4& 90.2 & 38.3 & 28.0 & 55.5 & 68.1 & 67.5& 90.3 &61.2 & 98.0 & 54.1 & 60.1 &82.0 & 27.4 & 67.0 & 59.1 & 50.9 & 55.4 & 62.6 \\
        DePT & 82.0 & 47.3 & 89.8& 37.5 & 26.9 & 53.0 & 66.2 & 65.8 & 89.6 & 60.8 & 97.7 & 52.8 & 59.5 & 81.6 & 26.5 & 66.5 & 57.9 & 49.5 & 53.2 & 61.3 \\
        VDP  & 83.0 & 48.3 & 90.2 & 38.2 & 28.0 & 55.5 & 68.2 &67.5 & 90.2 & 61.2 & 97.9 & 54.1 & 59.9 & 82.0 & 27.5 & 67.0 & 59.1 & 50.9 & 55.3 & 62.3 \\
        \cellcolor{lightgray}\textbf{SVDP} &\cellcolor{lightgray}\textbf{85.2} &\cellcolor{lightgray}\textbf{54.4}&\cellcolor{lightgray}\textbf{91.1}&\cellcolor{lightgray}\textbf{43.1}&\cellcolor{lightgray}\textbf{31.8}&\cellcolor{lightgray}\textbf{57.7}&\cellcolor{lightgray}\textbf{69.2}&\cellcolor{lightgray}\textbf{69.9}&\cellcolor{lightgray}\textbf{90.8}&\cellcolor{lightgray}\textbf{62.1}&\cellcolor{lightgray}\textbf{98.2}&\cellcolor{lightgray}\textbf{56.5}&\cellcolor{lightgray}\textbf{60.7} 
 &\cellcolor{lightgray}\textbf{83.0}&\cellcolor{lightgray}\textbf{28.7}&\cellcolor{lightgray}\textbf{67.6}&\cellcolor{lightgray}\textbf{63.6}&\cellcolor{lightgray}\textbf{55.0}      &\cellcolor{lightgray}\textbf{55.7} 
 &\cellcolor{lightgray}\textbf{64.4}\\\bottomrule

        \Xhline{1.2pt}
    \end{tabular}
  }
\label{table:rain acdc}
\end{table*}

\begin{table*}[!htb]
\caption{Performance Comparison for \textbf{Cityscapes-to-ACDC Snow domain in TTA scenario}. The IoU score of each class and the mIoU score are reported. The best results are highlighted in \textbf{bold}.
  } 
  \centering
  \resizebox{0.99\textwidth}{!}{
    \def\arraystretch{1.1}
    \begin{tabular}{ l | c c c c c c c c c c c c c c c c c c c | c }
        \Xhline{1.2pt}
        Method & \rotatebox{0}{road} & \rotatebox{0}{side.} & \rotatebox{0}{buil.} & \rotatebox{0}{wall} & \rotatebox{0}{fence} & \rotatebox{0}{pole} & \rotatebox{0}{light} & \rotatebox{0}{sign} & \rotatebox{0}{veg.} & \rotatebox{0}{terr.} & \rotatebox{0}{sky} & \rotatebox{0}{pers.} & \rotatebox{0}{rider} & \rotatebox{0}{car}& \rotatebox{0}{truck} & \rotatebox{0}{bus} & \rotatebox{0}{train} & \rotatebox{0}{mbike} & \rotatebox{0}{bike} & mIoU \\
        \hline
        \hline
        Source  & 79.8 & 40.8 & 86.9 & 43.6 & 46.5 & 56.4 & 72.3 & 65.5 & 82.9 &  \textbf{5.7} & 97.1 & 62.8 & 40.4 & 85.4 & 54.7 & 44.5 & 73.1 & 22.7 & 36.0 & 57.8 \\
        TENT & 79.6 & 40.0 & 86.8 & 43.4 & 46.5 & 56.1 & 72.2 & 65.6 & 82.9 &  \textbf{5.7} & 97.1 & 63.0 & 40.9 & 85.5 & 54.7 & 43.1 & 72.7& 23.0 & 36.5& 57.7\\
        CoTTA & 80.1 & 40.7 & 87.5 & 43.9 & 47.7 & 59.9 & 75.3 & 69.2 & 84.0 & 5.1 & 97.2 & 67.3 & 46.9 & 86.2 & 56.1& 43.4 & 74.1 &  25.7& 43.3 & 59.8 \\
        DePT & 79.1 & 40.6 & 86.8 & 43.4 & 47.5 & 59.8 & 75.1 & 69.4 & 83.5 & 5.2 & 97.1 & 67.2 & 46.5 & 86.3 & 56.0 & 44.0 & 73.9 & 25.6 & 43.1 & 59.5 \\
        VDP & 80.1 & 40.8 & 87.5 & 43.9 & 47.8 & 59.9 & 75.1 & 69.4 & 83.9 & 5.1 & 97.2 & 67.2 & 46.7 & 86.2 & 56.2 & 43.9 & 74.0 & 25.7 & 42.9 & 59.7 \\
        \cellcolor{lightgray}\textbf{SVDP} &\cellcolor{lightgray}\textbf{86.6} &\cellcolor{lightgray}\textbf{57.7}&\cellcolor{lightgray}\textbf{88.2}&\cellcolor{lightgray}\textbf{47.0}&\cellcolor{lightgray}\textbf{45.5}&\cellcolor{lightgray}\textbf{62.4}&\cellcolor{lightgray}\textbf{75.9}&\cellcolor{lightgray}\textbf{71.5}&\cellcolor{lightgray}\textbf{84.9}&\cellcolor{lightgray}3.9&\cellcolor{lightgray}\textbf{97.4}&\cellcolor{lightgray}\textbf{68.0}&\cellcolor{lightgray}\textbf{49.3} 
 &\cellcolor{lightgray}\textbf{87.1}&\cellcolor{lightgray}\textbf{57.1}&\cellcolor{lightgray}\textbf{49.2}&\cellcolor{lightgray}\textbf{77.5}&\cellcolor{lightgray}\textbf{26.8 }     &\cellcolor{lightgray}\textbf{45.4} 
 &\cellcolor{lightgray}\textbf{62.2}\\\bottomrule
        \Xhline{1.2pt}
    \end{tabular}
  }
\label{table:snow acdc}
\end{table*}

\section{Additional Quantitative Results}
\label{Sec: quantitative}
We present a comprehensive presentation of experimental results on the Test-time adaptation task for Cityscapes-to-ACDC, as shown in Tab .~\ref{table:fog acdc} - Tab .~\ref{table:snow acdc}. Our findings suggest that our proposed approach can better address the domain shift problem and achieve better IoU value in most categories.

\section{Additional related work}
\label{Sec:related}

\textbf{Semantic segmentation} is a crucial task in many computer vision applications aimed at assigning a categorical label to every pixel in an image. Several representative works in this field include DeepLab \cite{chen2017deeplab}, PSPNet \cite{zhao2017pyramid}, RefineNet \cite{lin2017refinenet}, and Segformer \cite{xie2021segformer}. Despite their high performance, these methods usually require extensive amounts of pixel-level annotated data, which can be laborious and time-consuming to collect. Additionally, they may suffer from poor generalization when applied to new domains.
Recent research has focused on addressing these challenges through domain adaptation strategies. For instance, \cite{yang2020fda} proposes a method that swaps the low-frequency spectrum to align the source and target domains. \cite{tranheden2021dacs} mixes the images from both domains, along with their corresponding labels and pseudo-labels. In contrast, \cite{wu2021dannet} uses adversarial learning to train a domain adaptation network for nighttime semantic segmentation. \cite{hoyer2022daformer} develops a novel model and training strategies to enhance training stability and avoid overfitting to the source domain.
However, these methods often require retraining the model on the source domain, which is inconvenient. Furthermore, they need to be retrained when adapting to a new target domain, incurring additional time and resource costs. Therefore, we propose SVDP to efficiently address the domain shift problem, which leverages a pre-trained model on the source domain and adds only a few parameters to achieve strong generalization capabilities on the target domain.

\textbf{Depth estimation} 
Depth estimation is a crucial task of machine scene comprehension. The ascendancy of deep learning has established it as the dominant approach for supervised depth estimation across both outdoor~\cite{eigen2014depth, geiger2012we, yang2019drivingstereo} and indoor settings~\cite{Silberman:ECCV12, scharstein2014high}. Typical methodologies involve the integration of a universal encoder, responsible for assimilating global context, alongside a decoder designed to retrieve depth details~\cite{Xu_2018_CVPR, Ramamonjisoa_2020_CVPR, lee2019monocular, Ramamonjisoa_2019_ICCV, fu2018deep}. In the context of cross-domain depth estimation, the emphasis is on aligning source and target domains either at the input or feature level~\cite{kundu2018adadepth,zheng2018t2net,zhao2019geometry}. As a case in point, the work presented in~\cite{zhao2019geometry} introduced a geometry-centric symmetric adaptation framework, conceived to optimize both the translation and the depth estimation processes simultaneously.~\cite{li2023test} combines the self-supervised model and the supervised model to tackle this problem. However, these methods always lead to catastrophic forgetting, and thus we introduce the Sparse Visual Domain Prompt to tackle this problem.
% \bibliography{supbib}

\end{document}